# BigDL 2.0: Seamless Scaling of AI Pipelines from Laptops to Distributed Cluster

Jason (Jinquan) Dai, Ding Ding, Dongjie Shi, Shengsheng Huang, Jiao Wang,
Xin Qiu, Kai Huang, Guoqiong Song, Yang Wang, Qiyuan Gong, Jiaming Song,
Shan Yu, Le Zheng, Yina Chen, Junwei Deng, Ge Song

Intel

## Abstract

*Most AI projects start with a Python notebook running on a single laptop; however, one usually needs to go through a mountain of pains to scale it to handle larger dataset (for both experimentation and production deployment). These usually entail many manual and error-prone steps for the data scientists to fully take advantage of the available hardware resources (e.g., SIMD instructions, multi-processing, quantization, memory allocation optimization, data partitioning, distributed computing, etc.). To address this challenge, we have open sourced BigDL 2.0 at* https://github.com/intel-analytics/BigDL/ *under Apache 2.0 license (combining the original BigDL [19] and Analytics Zoo [18] projects); using BigDL 2.0, users can simply build conventional Python notebooks on their laptops (with possible AutoML support), which can then be transparently accelerated on a single node (with up-to 9.6x speedup in our experiments), and seamlessly scaled out to a large cluster (across several hundreds servers in real-world use cases). BigDL 2.0 has already been adopted by many real-world users (such as Mastercard, Burger King, Inspur, etc.) in production.*

## 1. Introduction

Applying AI models to end-to-end data analysis pipeline plays a critical role in today's large-scale, intelligent applications. On the other hand, AI projects usually start with a Python notebook running on a single laptop or workstation, and one needs to go through a mountain of pains to scale it to handle larger dataset with high performance (for both large-scale experimentation and production deployment). These often require the data scientists to follow many manual, error-prone steps and even to make intrusive code changes, so as to fully take advantage of the available hardware resources (e.g., SIMD instructions [26], multi-processing [31] [27], quantization [25], memory allocation optimization [9] [21], data partitioning, distributed computing [39] [34] [20] [28], etc.).

To address these challenges, we have open sourced BigDL 2.0 at https://github.com/intel-analytics/BigDL/ under Apache 2.0 license (combining the original BigDL [19] and Analytics Zoo [18] projects), which allows users to build end-to-end AI pipelines that are transparently accelerated on a single node (with up-to 9.6x speedup in our experiments) and seamlessly scaled out to a large cluster (across several hundreds of nodes in real-world use cases). To make it easy for the data scientists to build large-scale, distributed AI applications, we have adopted the following design principles.

- **Standard APIs.** Using BigDL 2.0, users can simply build conventional Python notebooks on their laptops using standard APIs (such as Tensorflow [10] or PyTorch [32]); all the tuning, accelerations and scaling-out are automatically handled by the underlying toolkit.

- **End-to-end pipeline.** The toolkit should take a holistic approach that optimizes the entire AI pipeline (from data preprocessing, feature transformation, hyperparameter tuning [37], model training and inference, model optimization [17] [25] and deployment, etc.).

- **Transparent acceleration.** The toolkit should help users transparently accelerate their AI pipelines for training or inference, by automatically integrating optimized libraries, best-known configurations, and software optimizations.

- **Seamless scaling.** The toolkit should seamlessly scale out the end-to-end AI pipelines (including distributed data-parallel processing, model training, tuning and inference) with simple and familiar APIs for the data scientists.

The rest of this paper is organized as follows. Sec. 2 presents the overall design of BigDL 2.0; Sec. 3 and Sec. 4 describes in details the architecture, functionalities and

APIs for transparent acceleration and seamless scaling respectively. Finally, Sec. 5 shares some real-world use cases of BigDL 2.0, and Sec. 6 concludes the paper.

## 2. BigDL 2.0

As described in Sec. 1, BigDL 2.0 combines the original BigDL [19] and Analytics Zoo [18] projects, and transparently accelerates and seamless scales the end-to-end AI pipeline. These goals are accomplished through two libraries in BigDL 2.0, namely, BigDL-Nano and BigDL-Orca.

- *BigDL-Nano.* Leveraging several optimization techniques, such as using SIMD instructions [26], multi-processing [27] [31], quantization [6], memory allocation optimization [9] [21], model optimizations [5], etc., we have observed up to 10x speedup that significantly reduce the time to the solution when developing AI pipelines. However, applying these techniques requires using different tools, following complex steps, making intrusive code changes, and tuning many configurations, which are complex, error-prone, and hard to maintain for data scientists. To address this problem, we have integrated these optimizations into BigDL-Nano, so that users can transparently accelerate their deep learning pipelines (with possible AutoML [11] [29]) on a local laptop or a single server.

- *BigDL-Orca.* When scaling AI applications from a local laptop to distributed clusters, a key challenge in practice is how to seamlessly integrate distributed data processing and AI programs into a single unified pipeline. BigDL-Orca automatically provisions Big Data and AI systems (such as Apache Spark [38] [2] and Ray [31] [8]) for the distributed execution; on top of the underlying systems, it efficiently implements the distributed, in-memory data pipelines (for Spark Dataframes [12], Ray Datasets [8], TensorFlow Dataset [10], PyTorch DataLoader [32], as well as arbitrary python libraries), and transparently scales out deep learning (such as TensorFlow and PyTorch) training and inference on the distributed dataset (through scikit-learn style APIs [15]).

## 3. Transparent Acceleration with BigDL-Nano

Fig. 1 illustrates the architecture of BigDL-Nano. It utilizes dozens of acceleration technologies and tools (such as hardware specific configurations, SIMD instructions [26], multi-processing [31] [27], memory allocation optimization [21] [9], graph optimization and quantization [5] [4]) at its backend, and transparently accelerates both the model training and inference pipelines. For each of the acceleration technologies and libraries, BigDL-Nano adaptively applies proper configurations based on the user's execution environment, dataset, and models. BigDL-Nano brings all these accelerations to user transparently, so as to relieve the data scientist from manually tuning various configurations, applying different tools, or even making intrusive code changes.

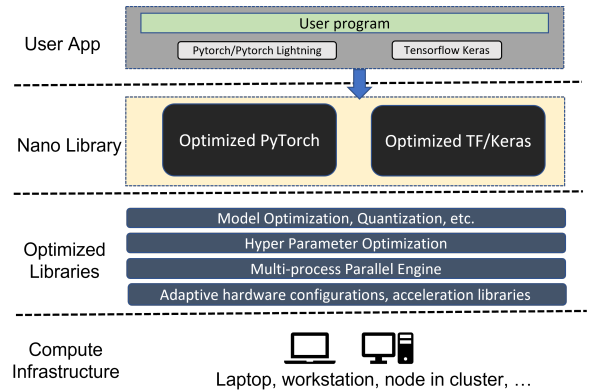

Figure 1. BigDL-Nano architecture

### 3.1. Accelerating Training Pipelines

To accelerate the end-to-end training pipelines, BigDL-Nano provides a transparent API that requires minimum changes in in user's original Tensorflow or PyTorch programs. For example, for PyTorch Lightning [22] users, normally they just need to change the library imports to use BigDL-Nano, as shown in Fig. 2. Under the hood, a set of training optimizations (e.g., ISA vectorization [26], improved memory allocation [21] [9], multi-processing [27] [31], optimizations in Intel Extension for PyTorch [5], etc.) are automatically enabled, which brings up-to 5.8x speedup as shown in Sec. 3.4.

### 3.2. Accelerating Inference Pipelines

BigDL-Nano also provides a set of lightweight APIs for accelerating the inference pipeline (such as model optimizations and quantization). Fig. 3 shows an example of how to enable quantization and ONNX Runtime [7] at inference stage using BigDL-Nano. By automatically integrating various optimization tools (including ONNX Runtime, INC [6], OpenVINO [4], etc.), it brings up-to 9.6x speedup as shown in Sec. 3.4.

### 3.3. AutoML

To optimize the model development productivity, BigDL-Nano also provides built-in AutoML [37] support through hyperparameter search. As shown in Fig. 4, by simply changing the import in the user program, BigDL-Nano collects the search spaces, passes them to the underlying HPO engine [13] [11] [29], and delays the instantiation of

```python
from bigdl.nano.pytorch.vision.datasets import
↪    ImageFolder
from bigdl.nano.pytorch.vision import transforms
from bigdl.nano.pytorch import Trainer
from torch.utils.data import DataLoader

data_transform = transforms.Compose([
   transforms.Resize(256),
   transforms.ColorJitter(),
   transforms.RandomCrop(224),
   transforms.RandomHorizontalFlip(),
   transforms.Resize(128),
   transforms.ToTensor()
])
dataset = ImageFolder(args.data_path,
↪    transform=data_transform)
train_loader = DataLoader(dataset,
↪    batch_size=batch_size, shuffle=True)
net = create_model(args.model, args.quantize)
trainer = Trainer(max_epochs=1)
trainer.fit(net, train_loader)
```

Figure 2. Example of PyTorch training using BigDL-Nano. Normally you just need to change the import from `pytorch_lightning import Trainer` to `from bigdl.nano.pytorch import Trainer`. And if you use torchvision, you also need to change the vision related imports. No changes are needed in other code.

```python
#... define the model Net
model = Net().to(device)
model.train()
#... omit the train loop here
# instantiate a trainer
trainer = bigdl.nano.pytorch.Trainer()
# use trainer to quantize the model and enable onnx
model_onnx_int8 = trainer.quantize(model,
    precision="int8",accelerator="onnxruntime", ...)
output = model_onnx_int8(data)
```

Figure 3. Example of PyTorch inference with quantization and onnx using BigDL-Nano. Only 2 extra lines of code are needed to enable quantization [6] and ONNX Runtime [7] at the same time.

the corresponding objects until the actual pipeline is configured and executed in each trial.

### 3.4. Performance Results

In this section we present some performance results of BigDL-Nano. In the tests, we run a classic image classification pipeline, i.e. classify cats and dogs using ResNet50 [3], on specific hardware platforms, and measured the end-to-end performance of training and inference with and without BigDL-Nano.

We have tested two scenarios using the same pipeline, i.e. *"train from scratch"* and *"transfer learning"*. In

```python
from bigdl.nano.tf.keras import Sequential
from bigdl.nano.tf.keras.layers import Dense,
↪    Flatten, Conv2D
import bigdl.nano.automl.hpo.space as space

model = Sequential()
model.add(Conv2D(
   filters=space.Categorical(32, 64),
   kernel_size=space.Categorical(3, 5),
   strides=2,
   activation="relu",
   input_shape=input_shape))
model.add(Flatten())
model.add(Dense(CLASSES, activation="softmax"))
...
model.compile(...)
model.search(...)
model.fit(...)
```

Figure 4. AutoML Example in BigDL-Nano. User can specify search spaces in layer arguments, etc. and then use `search` to search for the best hyperparameters.

"train from scratch" scenario, weights of all layers can be updated in the training stage, while in "transfer learning" scenario, only some of the layers can be updated and others are frozen. In the "transfer learning" scenario, BigDL-Nano brings bigger speedup (up to 5.8x) in the training stage due to frozen layers; the acceleration speedup stays the same (up to 9.6x) in inference stage for both scenarios, as optimizations (e.g. ONNX Runtime [7] and quantization [6]) are applied on all layers for inference.

For test platforms, we have chosen a ***laptop*** and a ***container*** reserved from a server. Although many data scientists use laptops for local experiments, it is also common for data scientists in organizations to use containerized environment allocated from remote server (e.g., on cloud). The detailed configuration of the laptop, and the container are as follows:

- **laptop** - a laptop with a single 8-core Intel(R) Core (TM) i7-11800H CPU @ 2.30GHz, 12G Memory, and OS is Ubuntu 20.04

- **container** - a docker [30] container with 28 cores in a single socket Intel(R) Xeon(R) Platinum 8380H CPU @ 2.90GHz, 192G memory, and OS is Ubuntu 16.04.

Fig. 5 and Fig. 6 summarized the speedup of training and inference using BigDL-Nano for two test scenarios on the laptop and on the container respectively. As the result shows, BigDL-Nano can bring up to 5.8x speedup in training and up to 9.6x speedup in inference in both platforms without accuracy lost. Powerful hardware with more cores (like container in cloud) could get higher acceleration rates than laptop in both training and inference; and int8 (using quantization) generally brings higher speed-up in inference than fp32 (using ONNX Runtime).

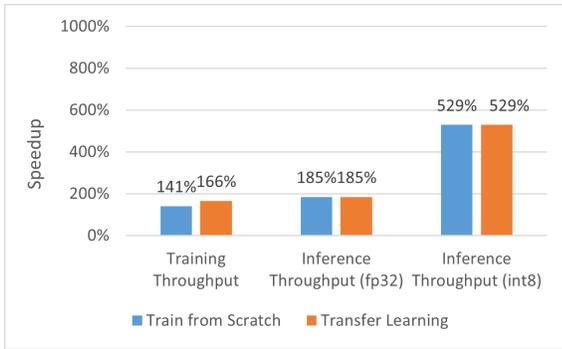

Figure 5. Speedup of BigDL-Nano for training and inference on Laptop

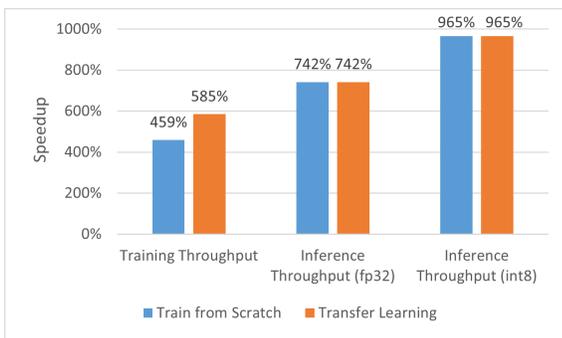

Figure 6. Speedup of BigDL-Nano and training and inference on Container

## 4. Seamless Scaling with BigDL-Orca

Fig. 7 shows the overall architecture of BigDL-Orca. To seamlessly scale the end-to-end AI pipelines from laptop to distributed cluster, BigDL-Orca will automatically provision Apache Spark [38] [2] and/or Ray [31] [8] as the underlying execution engine for the distributed data processing and model training/inference. On top of the distributed engine, the user can simply build his or her data pipeline in a data-parallel fashion (using TensorFlow Dataset [10], PyTorch DataLoader [32], Spark Dataframes [12], Ray Datasets [8], as well as arbitrary Python libraries such as OpenCV [14], Pandas [36], SciPy [35], spaCy [24], and etc.); then within the same program, the user can use sklearn-style Estimator APIs [15] in BigDL-Orca to directly apply AI models (such as Tensorflow [10], PyTorch [32], MXNet [16], etc.) to the processed data for distributed training and inference.

### 4.1. Distributed Data Processing Pipeline

BigDL-Orca supports four types of distributed data processing, namely, *TensorFlow Dataset or PyTorch DataLoader*, *Spark Dataframe*, *Ray Dataset*, and *XShards* (for arbitrary Python libraries).

**Tensorflow Dataset or PyTorch DataLoader.** User can directly use standard Tensorflow Dataset or PyTorch DataLoader to build their data processing pipeline, just as they do in single-node Tensorflow or PyTorch program, which can then be directly used for distributed deep learning training or inference, as shown in Fig. 8. Under the hood, BigDL-Orca automatically replicates the TensorFlow Dataset or PyTorch DataLoader pipeline on each node in the cluster, shards the input data, and executes the data pipelines using Apache Spark and/or Ray in a data-parallel fashion.

**Spark DataFrame or Ray Dataset.** Spark DataFrame is a common distributed data structure which allows users to apply various transformations on large-scale distributed data. Ray Dataset is a distributed data structure in Ray [31] [8] which is the standard way to load and exchange data in Ray libraries and applications. Both Spark DataFrame and Ray Dataset can be directly used for TensorFlow/PyTorch training or inference without data conversion when using BigDL-Orca, as shown in Fig. 9.

**XShards (for arbitrary Python libraries).** The XShards API in BigDL-Orca allows the user to process large-scale dataset using existing Python codes in a distributed and data-parallel fashion. When scaling a local AI pipeline to distributed cluster, a major challenge for the users is to rewrite their data ingestion or processing codes so as to support distributed data storage or structure (e.g., using new distributed data processing libraries). Such code modification requires the user to learn new APIs, and is error-prone when there is inconsistency between the user code and new libraries.

Using XShards, the users can enable distributed data loading and transformation by reorganizing – instead of rewriting – the original Python code, as illustrated in Fig. 10. In essence, an **XShards** contains an automatically sharded (or partitioned) Python object (e.g., Pandas [36] Dataframe, Numpy [23] NDArray, Python Dictionary or List, etc.). Each partition of the XShards stores a subset of the Python object and is distributed across different nodes in the cluster; and the user may run arbitrary Python codes on each partition in a data-parallel fashion using XShards.transform_shard.

### 4.2. Distributed Training and Inference Pipeline

BigDL-Orca provides sklearn-style APIs (namely, **Estimator**) for transparently distributed model training and inference. To perform distributed training and inference, the user can first create an BigDL-Orca Estimator

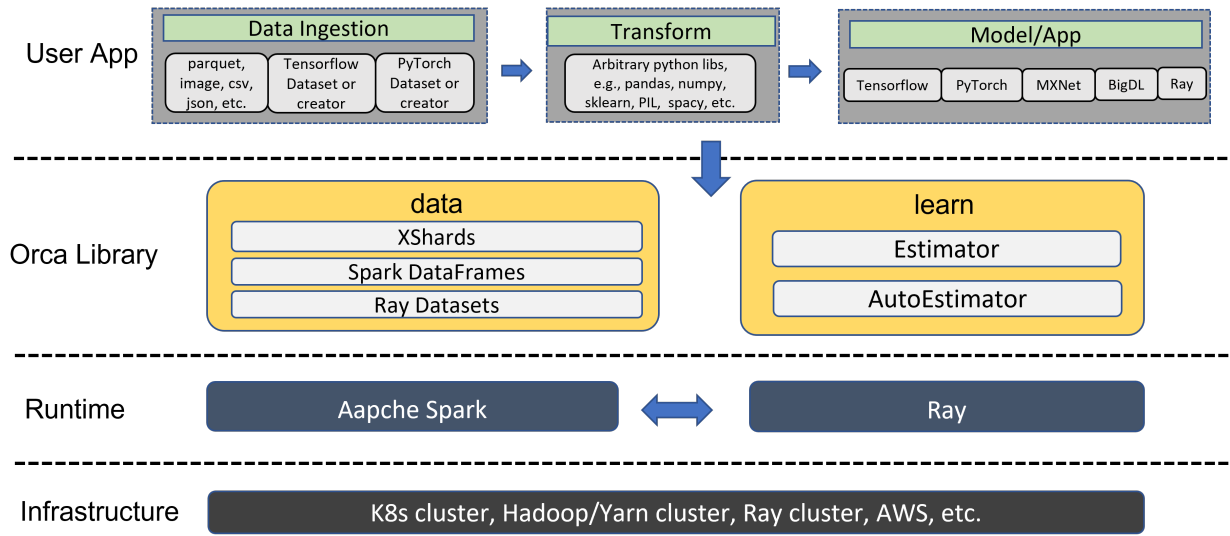

Figure 7. BigDL-Orca architecture

```
import torch
from torchvision import datasets, transforms
from bigdl.orca.learn.pytorch import Estimator

train_loader = torch.utils.data.DataLoader(
        datasets.MNIST(...),
        batch_size=batch_size,
        shuffle=True)

est = Estimator.from_torch(model=torch_model,
    optimizer=torch_optim, loss=torch_criterion)
est.fit(data=train_loader)
```

Figure 8. Example of using PyTorch loaders as input of Estimator

```
df = spark.read.parquet("data.parquet") # df is a Spark
    DataFrame
est = Estimator.from_keras(keras_model=model)
est.fit(data=df,
        feature_cols=['user', 'item'], # specifies
            which column(s) to be used as inputs
        label_cols=['label']) # specifies which
            column(s) to be used as labels
```

Figure 9. Estimator takes Spark DataFrame directly as input

from any standard (single-node) TensorFlow, Keras or PyTorch model, and then call `Estimator.fit` or `Estimator.predict` methods (using the data-parallel processing pipeline as input), as illustrated in Fig. 11.

Under the hood, the BigDL-Orca Estimator will replicate the model on each node in the cluster, feed the data partition (generated by the data-parallel processing pipeline) on each node to the local model replica, and

```
def negative(df, column_name):
    df[column_name] = df[column_name] * (-1) # pandas
        code
    return df

train_shards = shards.transform_shard(negative,
    'value')
```

Figure 10. Example of using XShards. The original transformation `df[column_name] = df[column_name] * (-1)` is reorganized into a function `negative` and applied on shard partitions using `transform_shard`.

```
#PySpark DataFrame
train_df = sqlcontext.read.parquet(...).withColumn(...)
test_df = ...
#TensorFlow Model
from tensorflow import keras
...
model = keras.Model(inputs=[user, item],
    outputs=outputs)
model.compile(optimizer= "adam",
              loss= "sparse_categorical_crossentropy",
              metrics=['accuracy'])

#Distributed training and inference using BigDL-Orca
est = Estimator.from_keras(keras_model=model)
est.fit(train_df, feature_cols=['user', 'item'],
    label_cols=['label'])
est.predict(test_df, feature_cols=['user', 'item'])
```

Figure 11. Example of using `Estimator` to train a Tensorflow keras model on PySpark Dataframe and do inference.

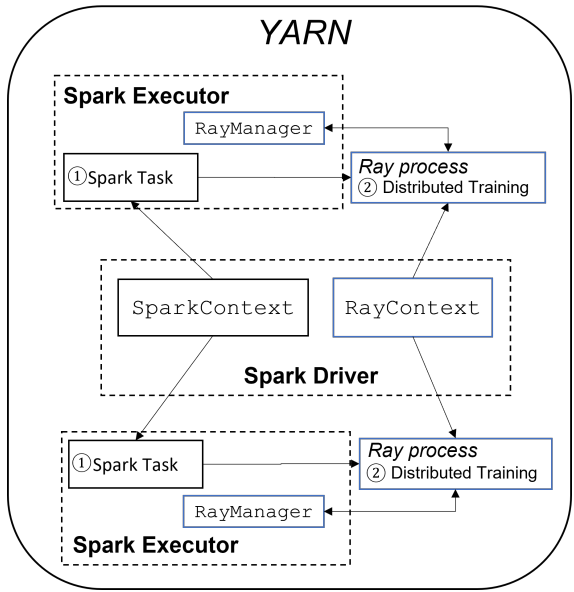

Figure 12. RayOnSpark Architecture

synchronize model parameters using various backend technologies (such as `torch.distributed` [27], `tf.distribute.MirroredStrategy`, Horovod [33], or the parameter sync layer in BigDL [19]).

Furthermore, BigDL-Orca also provides an **`AutoEstimator`** API, which can help users to train models with built-in support for automatic hyperparameter tuning [29] [11]. `AutoEstimator` has a similar interface as `Estimator`, with extra tuning related configurations such as search space and search algorithms, etc.

### 4.3. Ray on Spark Pipeline

In addition to distributed deep learning training and inference, BigDL-Orca also seamlessly integrates Ray [8] into Big Data platform through the *RayOnSpark* support. Fig. 12 illustrates the architecture of RayOnSpark. A `SparkContext` is first created on the driver responsible for launching multiple Spark executors; in RayOnSpark, the Spark [38] driver program also creates a `RayContext` to automatically launch Ray processes alongside each Spark executor. RayOnSpark will also create a `RayManager` inside each Spark executor to manage Ray processes (e.g., automatically shutting them down when the training finishes). As a result, the user can directly write Ray code inside the Spark program, which allows Ray applications to be seamlessly integrated into Big Data processing pipeline and directly run on in-memory Spark RDDs or DataFrames.

## 5. Real-World Use Cases

In this section, we share some real-world use cases of BigDL 2.0 at Mastercard and Inspur.

**"AI at Scale" at Mastercard.** Building on top of BigDL 2.0, Mastercard has adopted an "AI at Scale" approach to accelerate the entire machine learning lifecycle (including data analysis, experimentation, model training, deployment, resource optimizations, monitoring, etc.) [1]. This is accomplished by building a unified Big Data AI architecture with BigDL 2.0 on hybrid data/ML infrastructures (which automates AI/ML pipelines and model lifecycle management). Consequently, Mastercard engineers are able to seamlessly integrated big data analysis (Spark ecosystem) and deep learning (using TensorFlow and Keras) into end-to-end AI applications, which seamlessly scale to distributed Intel Xeon clusters for distributed training and serving. As a result, Mastercard can avoid the additional cost and complexity of special-purpose processors, while their AI training jobs can complete within only 5 hours on average (running on several hundred Intel Xeon servers to support up to 2.2 billion users and hundreds of billions of records).

**Smart Transportation Solution at Inspur.** Inspur have built their end-to-end, CV (computer-vision) based Smart Transportation solution using BigDL 2.0. The solution provides a unified big data and AI analysis platform integrated with big data preprocessing, model training, inference, and existing big data processing workflows. In particular, it builds the end-to-end pipeline from distributed video stream data processing, to distributed AI model training/inference (including multi-object tracking and OCR), and to vehicle trajectory binding, with 30% higher performance and 40% lower cost. We invite the readers to refer to the previous CVPR 2021 tutorial session [18] for more details.

## 6. Conclusion

In this paper we presented *BigDL 2.0*, an open source Big Data AI toolkit (https://github.com/intel-analytics/BigDL/). Using BigDL 2.0, users can simply build conventional Python notebooks on their laptops (with possible AutoML support), which can be transparently accelerated on a single node (with up-to 9.6x speedup as shown in our experiments), and seamlessly scaled out to a large cluster (across several hundreds servers as shown in real-world use cases). BigDL 2.0 has already been adopted by many real-world users (such as Mastercard, Burger King, Inspur, etc.) in production.